\documentclass{article}

\usepackage{arxiv}
\usepackage[english]{babel}
\usepackage{verbatim}
\usepackage{titlesec}
\usepackage{cite} 
\usepackage{comment}
\usepackage{url}
\usepackage{xcolor}
\usepackage{multirow, soul, amsmath}
\usepackage{graphicx}
\usepackage{tikz}
\usepackage{pdflscape}
\usepackage{xltabular} 
\usepackage{ragged2e}  
\newcolumntype{L}{>{\RaggedRight\hspace{0pt}}X}
\usepackage{enumitem}
\usepackage{booktabs}
\usepackage{pifont}
\usepackage{hyperref}
\usepackage{lscape}
\usepackage{multicol,multirow}
\newcommand{\greencheck}{\textcolor{green!80!black}{\ding{51}}}
\newcommand{\crossmark}{\textcolor{red}{\ding{55}}}

\newlist{mylist}{enumerate*}{1}
\setlist[mylist]{label=(\roman*)}

\title{A Review on Generative AI Models for Synthetic Medical Text, Time Series, and Longitudinal Data\thanks{\textbf{This paper is currently under review by NPJ Digital Medicine.}}}

\author{Mohammad~Loni$^1$, Fatemeh~Poursalim$^2$, Mehdi~Asadi$^3$, Arash~Gharehbaghi$^4$}

\date{
    $^1$School of Innovation, Design and Engineering,
Mälardalen University, Sweden \\ \texttt{mohammad.loni@mdu.se}\\[2ex]%      %$^2$Shiraz University of Medical Science, Shiraz, Iran  \\ \texttt{fpoursalim72@gmail.com}\\[2ex]%
     $^2$Servicehälsan Familjeläkare i Västerås AB, Västerås, Sweden \\ \texttt{fatemeh.poursalim@servicehalsan.se}\\[2ex]%
    $^3$ICT Department, Turku University of Applied Sciences, Turku, Finland \\ \texttt{mehdi.asadi@turkuamk.fi}\\[2ex]%
    $^4$Department of Biomedical Engineering, Link\"oping University, Link\"oping, Sweden \\ \texttt{arash.gharehbaghi@liu.se}\\%
}
     
\begin{document}
\maketitle

\begin{abstract}

This paper presents the results of a novel scoping review on the practical models for generating three different types of synthetic health records (SHRs): medical text, time series, and longitudinal data. The innovative aspects of the review, which incorporate study objectives, data modality, and research methodology of the reviewed studies, uncover the importance and the scope of the topic for the digital medicine context. In total, 52 publications met the eligibility criteria for generating medical time series (22), longitudinal data (17), and medical text (13). Privacy preservation was found to be the main research objective of the studied papers, along with class imbalance, data scarcity, and data imputation as the other objectives. The adversarial network-based, probabilistic, and large language models exhibited superiority for generating synthetic longitudinal data, time series, and medical texts, respectively. Finding a reliable performance measure to quantify SHR re-identification risk is the major research gap of the topic.

\end{abstract}

\section{Introduction}
\label{src:introduction}

Application of digital technologies, e.g. artificial intelligence, to improve medical management, patient outcomes, and healthcare delivery, is known as digital medicine \cite{Rajpurkar2022, McGenity2024}. This context has significantly evolved towards intelligent decision-making after the development of deep learning (DL) models. A common feature of most DL models is the need for a large dataset for training and validation of the models. Preparing a large dataset that incorporates sufficient samples of various classes for training a DL model is sometimes problematic. This is particularly evident when it comes to medical data in which the privacy of patient data demands serious attention. 

Social and medical communities rigorously assign escalating regulations at the different societal levels to avoid misconduct of patient data. In this light, the European Union has recently released the first regulation \cite{AIAct} on artificial intelligence towards lowering the risk of data abuse and further governance on the developers. Such pertinent restrictions as well as difficulties in collecting medical data, act as the impeding factors in preparing a sufficiently large and fair dataset for training and validation of the DL models. Medical data is nowadays stored in a digital fashion, named Electronic Health Record (EHR), which is composed of a set of the health-related data collected from the individuals of a population during their visits to any care unit defined by the healthcare system of the population \cite{ghosheh2024survey}. An EHR typically contains demographic data, clinical findings, lab values, procedures, medications, symptoms, diagnoses, medical images, physiological signals, and descriptive texts obtained from the individuals during the visits. During a visit to a care unit, and depending on the visit, an individual may undergo a sequence of investigations and/or examinations, called events, which are stored in the EHR of the individual, using the globally known codes, e.g. international classification code (ICD) \cite{khoury2020international} for the diagnosis. A sequence of the tabular health records of an individual, resulting from several visits, constitutes longitudinal health data \cite{li2023generating, mosquera2023method}. 

Accessibility of the EHR can be of vital importance when it comes to the patient management, and hence, highest level of security considerations are administered to preserve privacy of EHRs. Healthcare systems set intensive restrictions and limitations for accessing EHR contents. As a result, preparing a rich clinical dataset is considered as an important research question for any study in the digital medicine context. A practical solution to this research question is to create a synthetic copy of the clinical dataset from the real one, and use it for research purposes. The created dataset, which is called synthetic health record (SHR), can be utilized and shared with the researchers, instead of the real ones, for the training and validation of DL models. A SHR is created to resemble the general characteristics of EHR data while ensuring the subjects of the EHR remain unidentifiable \cite{kaur2021application, nikolentzos2023synthetic, murtaza2023synthetic}. Such datasets can be employed by any machine learning method for training and optimization purposes. 

Health records can incorporate tabular data of the medical records, longitudinal tables of different visits, time series of physiological signals, textual clinical data, and medical images. Recent progress in developing DL models for generating various modalities of synthetic data enabled researchers to address the diverse research gaps that existed in the preparation of high-quality health data. Topics such as privacy leakage, limited data availability, uneven class distribution, and scattered data, are regarded as some of the research topics of this domain.

Despite substantial progress in developing generative models for synthetic medical data, there is still a big room for studying the reliability of the generated data. As we will see in the discussion section, the reliability of the models is explored from two different perspectives: the quality of generated data in terms of conformity with real data, and the capability of learning models to preserve the privacy of the real data in terms of re-identification. Discrepancy in the study objectives for various data modalities, and inconsistency in the evaluation metrics, make it a complicated task to select an appropriate model for creating SHR along with a realistic evaluation of the model.

This paper presents the results of a scoping review study on the existing DL models for generating SHRs. The review examines the DL models deployed, the modalities utilized, the datasets invoked, and the metrics employed by the researchers to explore the scope and potential of using SHR for different medical objectives. The study taxonomy is performed in both technical and applicative manners to represent the relevance of the models in conjunction with their capabilities in producing time series, text, and longitudinal SHR.
The main study objectives are the identification of the practical capabilities and the knowledge gaps of generative models for creating SHR, which are detailed as follows: 
\begin{itemize}[noitemsep]
    \item Finding state-of-the-art of the generative models for creating synthetic medical texts, time series, and longitudinal data, along with the methodological limitations.
    \item Summarizing the existing performance measures in conjunction with the related metrics for evaluating the quality of SHR.
    \item Listing the most used datasets employed by the researchers for generating SHR.
    \item Finding the key research gaps of the field. 
\end{itemize}

The unique features of this study are mainly the comprehensiveness of the review and the novel research taxonomy. As we will see in the discussion section, this paper introduces innovative features compared to the other review papers. These features initiate the following contributions to the field: 
\begin{itemize}[noitemsep]
    \item Introducing taxonomic novelty by defining various data modalities and applications. There are review papers on SHR for tabular and image data, however, less attention was paid to important topics such as medical texts and longitudinal data.
    \item Evaluation of different machine learning (ML) methods for generating various forms of SHR.
    \item Representing the performance measures and the evaluation metrics used for validating the quality of SHR in association with the related methodological capabilities for evaluating different data modalities. 
    \item Introducing available datasets for generating SHR in conjunction with the related applicability per the study taxonomy. 
\end{itemize}

It is worth noting that the number of publications in the SHR domain has drastically increased in the last two years. The novel aspects of this study project the scope and the possibilities of state-of-the-art methods in this new domain of digital medicine. 

\section{Background}
\label{src:background}

\subsection{Electronic Health Record (EHR)}
\label{src:background:EHR}

An EHR is composed of a set of health-related data collected from the individuals of a population during their visits to care units defined by the healthcare system of the population \cite{ghosheh2024survey}. The health-related data of an individual contains demographic data, clinical findings, lab values, procedures, medications, symptoms, diagnoses, medical images, physiological signals, and descriptive texts obtained from the individuals during the visits. The health information is holistically collected from an individual during each visit and includes temporal information such as dates and times. 

During a visit to a care unit, and depending on the visit, an individual may undergo a sequence of investigations and/or examinations, called events, which are stored in the EHR of the individual, using the globally known codes, e.g. international classification code (ICD\footnote{\url{https://www.who.int/standards/classifications/classification-of-diseases}}) for the diagnosis. Contents of a certain visit data are stored in tabular form, encompassing ICD codes in conjunction with the lab values, procedures, medications, and other medical data, collected throughout events that occurred during the visit. It is obvious that the events are patient-specific, and evermore, visit-specific, and may not follow a consistent sequence. A sequence of the tabular health records of an individual, resulting from several visits, constitutes longitudinal health data \cite{li2023generating, mosquera2023method}. 

It is important to note that a longitudinal health record  is not intrinsically regarded as a time series, for two main reasons: \begin{mylist}
    \item The data samples of a time series are collected in a priori known time stumps, in contrast with the longitudinal data in which the temporal information of the visits is not priori known, and
    \item The data samples of a time series exhibit a consistent dimension and structure, while the tabular contents of longitudinal data are strictly inconsistent and depend on the events that occurred during the visits.
\end{mylist} Nevertheless, longitudinal data can contain pointers to the time series of physiological measurements, e.g. electrocardiogram, to the images, e.g. Magnetic Resonance Imaging, and to the text prepared by the medical staff. The text data is attributed to the EHR of an individual at each visit by the authorized medical staff,  which reflects a summary of the findings after the visit. The texts are invoked by expert physicians for medical assessments and decision-making.

\subsection{Synthetic Health Record (SHR)}
\label{src:background:SHR}
A dataset of SHR is a synthetic counterpart to EHR, created to resemble the general characteristics of EHR data while ensuring the subjects of the EHR remain unidentifiable \cite{kaur2021application, nikolentzos2023synthetic, murtaza2023synthetic}. Such datasets can be employed by any machine learning method for training and optimization purposes. A SHR is supposed to preserve data distribution of the imitated EHR, and equally to be utilized by ML methods for training. This feature addresses the utility of a SHR. Meanwhile, the risk of identification of the subject personal data must be minimal, addressing re-identification feature of the SHR \cite{reiter2014bayesian, jordon2021hide, libbi2021generating}.

\subsection{Generative Models}
\label{src:background:method}

The existing methods for generating synthetic data typically fall into two categories: probability distribution techniques and neural network-based methods \cite{ghosheh2024survey}. Probability distribution techniques involve estimating a probability distribution of the real data and then drawing random samples that fit such distribution as synthetic data. Generative Markov-Bayesian probabilistic modeling is a technique used for synthesizing longitudinal EHRs \cite{wang2022using}. On the other hand, Recent developments in synthetic data generation are adopting advanced neural networks. Below are the most commonly used neural network-driven generative models:

\begin{itemize}[noitemsep]
    \item \textbf{Generative adversarial networks (GANs)} \cite{ghosheh2024survey}, comprising a generator and a discriminator, produce synthetic data resembling real samples drawn from a specific distribution. The discriminator distinguishes between real and synthetic samples, refining the generator's ability to create realistic data through adversarial training, enabling accurate approximation of the data distribution and generation of high-fidelity novel samples. GANs can generate sequences of data points that closely resemble the patterns observed in the original image or time series data.
    
    \item \textbf{Diffusion models} \cite{koo2023survey} gradually introduce noise into original data until it matches a predefined distribution. The core idea behind diffusion models is to learn the process of reversing this diffusion process, allowing for the generation of synthetic samples that closely resemble the original data while capturing its essential characteristics and variability. 

    \item  \textbf{Variational auto-encoders (VAEs)} \cite{ghosheh2024survey} are a category of generative models that learn to encode and decode data points while approximating a probability distribution, typically Gaussian, in the latent space. VAEs are trained by optimizing a variational lower bound on the log-likelihood of the data, enabling them to learn meaningful representations and create new data samples.

    \item \textbf{Large language models (LLMs)} \cite{peng2023study} can predict the probability of the next word in a text sequence based on preceding words, typically leveraging transformer architectures adept at capturing long-range dependencies. LLMs are effective for generating contextually appropriate text by learning the probability distribution of natural language data on vast corpora of text.
\end{itemize}

\subsection{Research Objectives of the Reviewed Publications}

The applicability of SHR was limited to providing input data for a better understanding of physiological systems in the early studies. On one hand, the development of data-intensive ML methods, and on the other hand, the restriction of sharing patient data, introduced by various data security legislators, e.g. GDPR \cite{GDPR2016a}, opened a new horizon to generative models for SHR. European Union has recently released the first regulation of artificial intelligence \cite{AIAct}, which introduces further restrictions to the field and researchers on using the research data to develop ML methods. The main objective of this topic may not be limited but can be summarized according to the literature review, as follows: 

\begin{itemize}[noitemsep]
 \item \textbf{Privacy:} Reliable SHRs can be generated based on patient data to be utilized for training and validation of ML methods.
 
 \item \textbf{Class Imbalance:} In many applications of health studies, access to different classes of data is not feasible in a consistent form, and a single class is dominantly seen in the study population. This can introduce bias to ML methods towards better learning of the dominant class. A reliable SHR can be invoked to generate synthetic data for minority classes.
 
\item \textbf{Data Scarcity:} Access to data of a specific class can sometimes be problematic. In this case, the scarce class is identified and modeled using the absolute minority samples along with meta-learning methods. The SHR is generated based on the identified model to explore the characteristics of the scarce class.

\item \textbf{Data Imputation:} EHR is heavily sparse with missing values, which are not uniformly obtained over the visits. Data imputation implies the methods to estimate the missing values that happened systematically or randomly in data collection.  
\end{itemize}

\section{Performance Measurement}
\label{src:performance_measurment}

Evaluating the strengths and weaknesses of generative models has become increasingly critical as these models continue to advance in complexity and capacity. The evaluation of generative models can be seen from different perspectives. In this study, we categorized evaluation metrics based on three different objectives: 
\begin{mylist}
    \item \textbf{Fidelity:} degree of faithfulness in which the synthetic data preserves the essential characteristics, structures, and statistical properties of the actual data. Fidelity can be seen in either population-based (e.g. examining marginal and joint feature distributions) or individual-based (e.g. synthetic data must adhere to specific criteria, such as not including prostate cancer in a female patient) levels, 
    \item \textbf{Re-identification:} concerning protection of sensitive information and confidentiality of individuals' identity, and 
    \item \textbf{Utility:} using synthetic data as a substitute for actual data for training/testing any medical devices and algorithms.
\end{mylist}

Evaluation techniques can also be categorized into two main groups: quantitative and qualitative methods: 
\begin{mylist} \item \textbf{Quantitative evaluation Methods:} Table~\ref{tab:evaluation_metrics} represents different quantitative evaluation metrics along with the evaluation objectives used to assess the effectiveness of generative models for the reviewed publications.

\item \textbf{Qualitative evaluation Methods:} Qualitative evaluation methods are often employed alongside quantitative measures to complement results with straightforward assessments. For instance, many studies utilize visualization approaches to compare distributions and embeddings of synthetic and real data, such as using histogram \cite{christensen2019fully_body, Asadi2023-mz}, Q–Q-plot \cite{Asadi2023-mz}, t-SNE \cite{yang2023ts, li2022tts, lee2020generating}, PCA \cite{li2022tts, yang2023ts}, and correlation \cite{christensen2019fully_body, kaur2021application}. Concerns regarding the clinical validity and trustworthiness of synthetic data pose significant obstacles to using it for clinical research. To tackle this issue, some studies have performed clinician evaluations, wherein medical professionals evaluate the realism of the synthetic data they are presented with \cite{Asadi2023-mz, peng2023study, wang2022using, wickramaratne2023sleepsim, alcaraz2023diffusion}. 
\end{mylist}

\begin{xltabular}{\textwidth}{b{45pt}b{100pt}b{40pt}b{30pt}b{40pt}b{120pt}}
\caption{A summary of the metrics used to evaluate generative models for creating SHRs.}
\label{tab:evaluation_metrics} 
\\ \toprule

\multirow{2}{*}{\textbf{Objective}} & \multirow{2}{*}{\textbf{Method}}  & \multicolumn{3}{c}{\textbf{Data Modality}} &  \multirow{3}{*}{\textbf{Ref. Study}}\\ \cmidrule{3-5} 

  &  &  \textbf{Time Series} & \textbf{Text} & \textbf{Longitud.} & \\  \midrule
\endfirsthead

\multicolumn{6}{c}%
{\tablename\ \thetable{} -- Continued from the previous page...} \\ \toprule 

\multirow{2}{*}{\textbf{Objective}} &  \multirow{2}{*}{\textbf{Method}} & \multicolumn{3}{c}{\textbf{Data Modality}} &  \multirow{3}{*}{\textbf{Ref. Study}} \\ \cmidrule{3-5} 

  &  &  \textbf{Time Series} & \textbf{Text} & \textbf{Longitud.} & \\ \midrule   
\endhead

\midrule \multicolumn{6}{r}{{Continued on the next page...}} \\ \endfoot

\endlastfoot

\parbox[t]{2mm}{\multirow{3}{*}{\rotatebox[origin=c]{90}{Fidelity}}} & Discriminative score \cite{ghosheh2024survey} & \multicolumn{1}{c}{\greencheck} & & \multicolumn{1}{c}{\greencheck} & \cite{li2023generating, wang2022wearable} \\ \cmidrule{2-6}
    
    &  Maximum mean discrepancy (MMD) \cite{el2022utility} & \multicolumn{1}{c}{\greencheck} & & \multicolumn{1}{c}{\greencheck} & \cite{yang2023ts, brophy2021multivariate, kiyasseh2020plethaugment, brophy2020synthesis, nikolaidis2019augmenting, li2023generating} \\ \cmidrule{2-6}
    
    & Multivariate Hellinger distance \cite{le2000asymptotics} & & & \multicolumn{1}{c}{\greencheck} & \cite{mosquera2023method} \\ \cmidrule{2-6}
    
    & Wasserstein distance \cite{el2022utility} & \multicolumn{1}{c}{\greencheck} & & \multicolumn{1}{c}{\greencheck} &  \cite{haleem2023deep, foomani2022synthesizing, shi2022generating, yoon2020anonymization} \\ \cmidrule{2-6} 

    & Inception score (IS) \cite{ghosheh2024survey} & \multicolumn{1}{c}{\greencheck} & & \multicolumn{1}{c}{\greencheck} & \cite{lee2021contextual, wang2020part} \\ \cmidrule{2-6}

    & Wilcoxon rank sum test \cite{murtaza2023synthetic} & \multicolumn{1}{c}{\greencheck} & & & \cite{maweu2021generating} \\ \cmidrule{2-6}
    
    & Kolmogorov–Smirnov (K-S) test \cite{ghosheh2024survey} & \multicolumn{1}{c}{\greencheck} & & \multicolumn{1}{c}{\greencheck} & \cite{haleem2023deep, Asadi2023-mz, foomani2022synthesizing, baowaly2019synthesizing, kaur2021application} \\ \cmidrule{2-6}

    & Euclidean distance (ED) \cite{murtaza2023synthetic} & \multicolumn{1}{c}{\greencheck} & & \multicolumn{1}{c}{\greencheck} & \cite{wang2022wearable, dahmen2019synsys, wang2020part, mosquera2023method}\\ \cmidrule{2-6}

    & Dimension-wise probability (DWPro) \cite{choi2017generating} & & & \multicolumn{1}{c}{\greencheck} & \cite{li2023generating, kaur2021application, li2023improving} \\ \cmidrule{2-6}

    & Dimension-wise prediction (DWPre) \cite{choi2017generating} & & & \multicolumn{1}{c}{\greencheck} & \cite{baowaly2019synthesizing, li2023improving} \\ \cmidrule{2-6}
    
    & Pairwise distance correlation \cite{murtaza2023synthetic} & \multicolumn{1}{c}{\greencheck} & & & \cite{haleem2023deep} \\ \cmidrule{2-6}

    & Pearson correlation \cite{hauke2011comparison} & & \multicolumn{1}{c}{\greencheck} & \multicolumn{1}{c}{\greencheck} & \cite{shi2022generating, li2023generating, peng2023study} \\ \cmidrule{2-6}

    & P-Value test & \multicolumn{1}{c}{\greencheck} & \multicolumn{1}{c}{\greencheck} & \multicolumn{1}{c}{\greencheck} & \cite{zhou2022datasifter, maweu2021generating, dash2020medical, shim2021synthetic} \\ \cmidrule{2-6}

    & Spearman rank correlation \cite{hauke2011comparison} & & & \multicolumn{1}{c}{\greencheck} & \cite{shi2022generating, wendland2022generation} \\ \cmidrule{2-6}

    & Kendall's rank correlation \cite{hauke2011comparison} & & & \multicolumn{1}{c}{\greencheck} & \cite{shi2022generating} \\ \cmidrule{2-6}

    & Kullback–Leibler (KL) -divergence \cite{ghosheh2024survey} & \multicolumn{1}{c}{\greencheck} & & \multicolumn{1}{c}{\greencheck} & \cite{lee2021contextual, wendland2022generation, zhang2021synteg} \\ \cmidrule{2-6}

    & Jensen–Shannon (JS) -divergence/-distance \cite{menendez1997jensen} & \multicolumn{1}{c}{\greencheck} & & \multicolumn{1}{c}{\greencheck} & \cite{haleem2023deep, li2022tts, zhang2022keeping, wendland2022generation, yoon2020anonymization} \\ \cmidrule{2-6}

    & Cosine similarity \cite{lahitani2016cosine} & \multicolumn{1}{c}{\greencheck} & \multicolumn{1}{c}{\greencheck} & & \cite{li2022tts, zhou2022datasiftertext} \\ \cmidrule{2-6}

     & Weighted latent difference \cite{ghosheh2024survey} & & & \multicolumn{1}{c}{\greencheck} & \cite{zhang2021synteg} \\ \cmidrule{2-6}

    &  Bilingual evaluation understudy (BLEU), self-BLEU \cite{papineni2002bleu} & & \multicolumn{1}{c}{\greencheck} & & \cite{hiebel2023can, zhou2022datasiftertext, kasthurirathne2021generative, al2021differentially, guan2019method, kasthurirathne2019adversorial, lee2018natural} \\  \cmidrule{2-6}

    & Jaccard similarity \cite{plattel2014distributed} &  & \multicolumn{1}{c}{\greencheck} & & \cite{al2021differentially} \\ \cmidrule{2-6}
    
    & $G^2$-Test \cite{rayson2004extending} & & \multicolumn{1}{c}{\greencheck} & & \cite{al2021differentially} \\ \cmidrule{2-6}
    
    & NLL-Test \cite{zhu2018texygen} (Likelihood-based) & & \multicolumn{1}{c}{\greencheck} & & \cite{al2021differentially, guan2019method} \\ \cmidrule{2-6}
    
    & ROUGE-N recall \cite{lin2004rouge} & & \multicolumn{1}{c}{\greencheck} & & \cite{libbi2021generating, lee2018natural} \\ \cmidrule{2-6}

    & N-grams overlap score \cite{hiebel2023can} & & \multicolumn{1}{c}{\greencheck} & & \cite{hiebel2023can} \\ \cmidrule{2-6}

    &  Consensus-based image description evaluation (CIDEr) \cite{bandi2023power} & & \multicolumn{1}{c}{\greencheck} & & \cite{lee2018natural} \\ \cmidrule{2-6}

    & Exact match difference \cite{bandi2023power} & & \multicolumn{1}{c}{\greencheck} & & \cite{shim2021synthetic} \\

\midrule

\parbox[t]{2mm}{\multirow{5}{*}{\rotatebox[origin=c]{90}{Re-identification}}} & Membership inference attack (MIA) \cite{ghosheh2024survey} & \multicolumn{1}{c}{\greencheck} & & \multicolumn{1}{c}{\greencheck} & \cite{zhang2021synteg, li2023generating, wang2022promptehr, brophy2021multivariate, brophy2020synthesis, theodorou2023synthesize} \\ \cmidrule{2-6} 

     & Attribute disclosure attack \cite{ghosheh2024survey} & & & \multicolumn{1}{c}{\greencheck} & \cite{kaur2021application, abell2021generating, zhang2021synteg, wang2022promptehr} \\ \cmidrule{2-6}
    
    & Differential privacy \cite{ghosheh2024survey} & \multicolumn{1}{c}{\greencheck} & \multicolumn{1}{c}{\greencheck} & \multicolumn{1}{c}{\greencheck} & \cite{wang2020part, li2023generating, kaur2021application, lee2020generating}  \\ \cmidrule{2-6}

     & Keywords inference attack & & \multicolumn{1}{c}{\greencheck} &  & \cite{wang2023differentially} \\ \cmidrule{2-6}
    
    & Bayesian disclosure attack \cite{reiter2014bayesian} & & & \multicolumn{1}{c}{\greencheck} & \cite{zhou2022datasifter} \\ \midrule

 \parbox[t]{2mm}{\multirow{5}{*}{\rotatebox[origin=c]{90}{Utility}}} & Sensitivity, Specificity \cite{ebrahimi2020review} & \multicolumn{1}{c}{\greencheck} & \multicolumn{1}{c}{\greencheck} & & \cite{kasthurirathne2019adversorial, maweu2021generating, kasthurirathne2021generative} \\ \cmidrule{2-6}

    & Precision \cite{ebrahimi2020review} & \multicolumn{1}{c}{\greencheck} & \multicolumn{1}{c}{\greencheck} & & \cite{libbi2021generating, yang2023ts, khademi2023data, peng2023study} \\ \cmidrule{2-6}

    & Recall \cite{ebrahimi2020review} & \multicolumn{1}{c}{\greencheck} & \multicolumn{1}{c}{\greencheck} & \multicolumn{1}{c}{\greencheck} & \cite{yang2023ts, khademi2023data, libbi2021generating, wang2022promptehr, peng2023study} \\ \cmidrule{2-6}

    & F1-Score \cite{ebrahimi2020review} & \multicolumn{1}{c}{\greencheck} & \multicolumn{1}{c}{\greencheck} & & \cite{yang2023ts, khademi2023data, kasthurirathne2021generative, shim2021synthetic, libbi2021generating, kasthurirathne2019adversorial, syed2019gender, lee2018natural, peng2023study, theodorou2023synthesize} \\ \cmidrule{2-6}
    
    & AUROC \cite{ebrahimi2020review} & \multicolumn{1}{c}{\greencheck} & \multicolumn{1}{c}{\greencheck} & \multicolumn{1}{c}{\greencheck} &  \cite{alcaraz2023diffusion, Asadi2023-mz, bing2022conditional, dash2020medical, kiyasseh2020plethaugment, kasthurirathne2021generative, kasthurirathne2019adversorial, syed2019gender, li2023improving, zhang2022keeping, wang2022promptehr, abell2021generating, zhang2021synteg, yoon2020anonymization, theodorou2023synthesize} \\ \cmidrule{2-6} 
    
    & Predictive Score \cite{yoon2019time} & \multicolumn{1}{c}{\greencheck} & & & \cite{wang2022wearable} \\ \cmidrule{2-6}
    
    & LPL and MPL \cite{wang2022promptehr} & & & \multicolumn{1}{c}{\greencheck} & \cite{wang2022promptehr} \\ \cmidrule{2-6}
    
    & Dynamic Time Warping (DTW) \cite{banko2012correlation} & \multicolumn{1}{c}{\greencheck} & & & \cite{wang2022wearable, brophy2020synthesis} \\ \cmidrule{2-6}

    & Multivariate Dynamic Time Warping (MVDTW) \cite{banko2012correlation} & \multicolumn{1}{c}{\greencheck} & & & \cite{dahmen2019synsys, brophy2021multivariate} \\ 

\bottomrule

\end{xltabular}

\section{Materials and Research Methodology}
\label{src:method}

\subsection{Study Taxonomy}
\label{src:method:taxonomy}

The study on SHRs covers a wide scope spanning from tabular medical data of a single record to longitudinal data from several visits and events. In contrast with the varieties in the adoption of different learning methods to create synthetic health data, the methodological suitability of the proposed methods depends merely on data modality. This review study is hence, performed according to the following taxonomy: 
\begin{mylist}
    \item longitudinal medical/health data,
    \item medical/health time series, and
    \item medical/health texts. 
\end{mylist}
 
\subsection{Research Method}
\label{src:method:search_method}

\begin{figure}[htbp]
    \centering
    \includegraphics[width=\columnwidth]{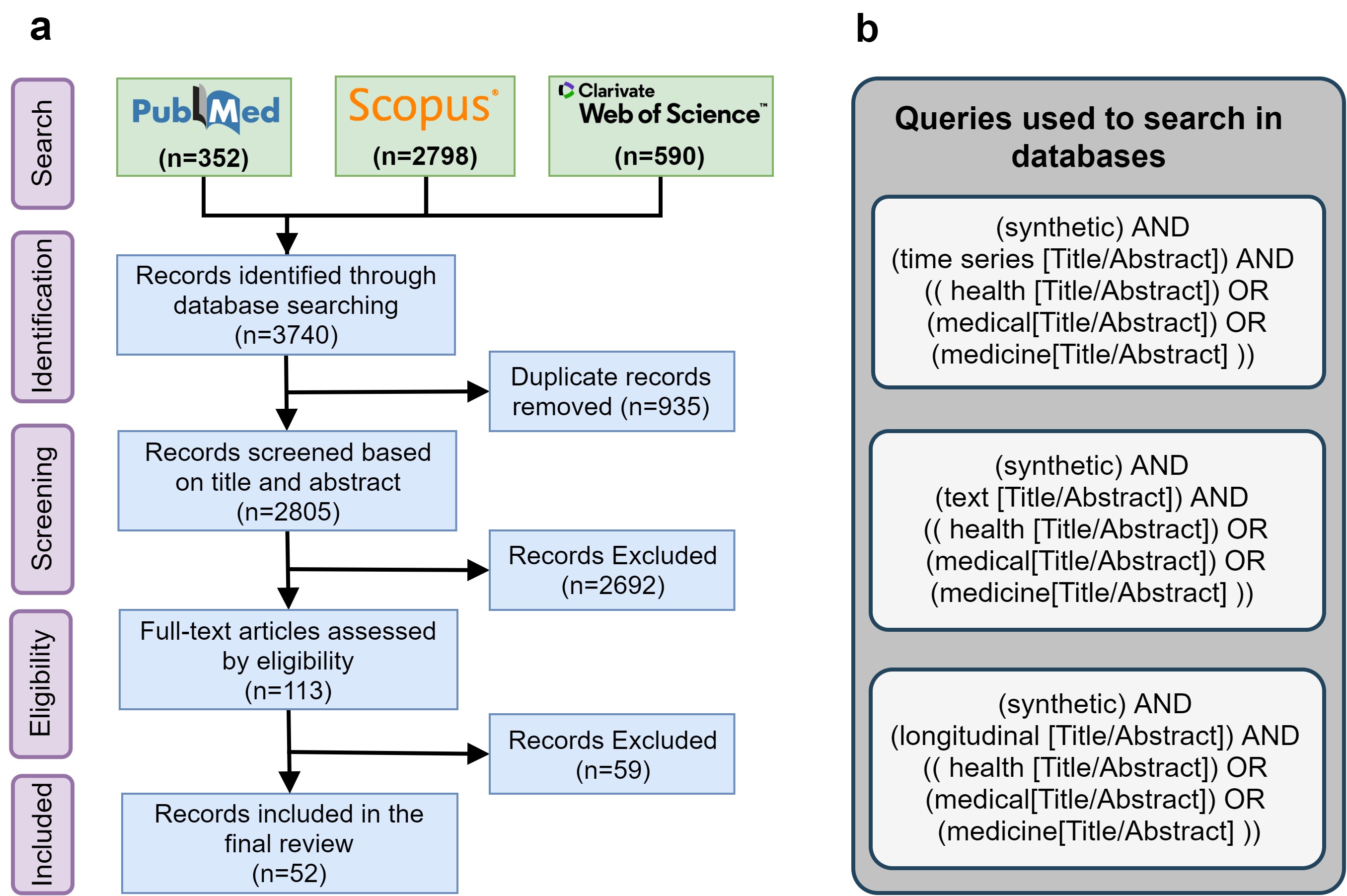}
    \caption{(a) Overview of the study selection process and (b) research queries used in the study.}
    \label{fig:flowchart}
\end{figure}

To perform this scoping review, we followed the recommendations outlined in the PRISMA-ScR guidelines \cite{tricco1967prisma}. The research method was comprised of 5 steps, described as follows: 

\begin{enumerate}[noitemsep]
    \item \textbf{Search:} A systematic search is performed on the three widely accepted platforms of scientific publications in this domain: PubMed, Web of Science, and Scopus using combinations of \{Synthetic\}, \{Time Series, Text, Longitudinal\}, and \{Medical, Medicine, Health\} keywords in the title and/or abstract of the publications. Our search queries are shown in Figure~\ref{fig:flowchart}. We adapted the string for each database, using various forms of the terms.
    
    \item \textbf{Identification:} The outcomes of the search were explored in terms of duplication and repeated publications were excluded from the study.
    
    \item \textbf{Screening:} In this phase Title and Abstract of the identified papers were studied and the topical relevance of the publications was investigated. Some of the publications from a different scientific topic were identified to participate in the study because of similarities in keywords. These publications were detected and excluded from the study.
    
    \item \textbf{Eligibility (inclusion criteria):} After the search phase, only those publications fulfilling all the below criteria were allowed to participate in the study:
\begin{mylist}
    \item published within 2018–2023, 
    \item the full paper is available, and 
    \item addressing an ML topic for electronic health record (EHR) generation.
\end{mylist}
Papers with only the Abstract available, cannot be analyzed and hence, excluded from the study, in addition to those addressing synthetic organs, without addressing the ML objectives.
    \item \textbf{Included Studies:} This study focuses on reproducible ML methods for generating synthetic, time series, longitudinal, or text contents of medical record. Therefore, the validity of the proposed methods in terms of implementation feasibility is an important criterion for consideration. We consolidated the scientific quality of the study by using the following Exclusion Criteria:
\begin{mylist}
    \item lack of the peer-reviewed process for the publication,
    \item EHR generation is not the major objective of the publication, and
    \item limited to tabular and image data only.
\end{mylist}
\end{enumerate}

\section{Results}
\label{src:results}

\subsection{Overview of the Findings}
\label{src:results:characteristics}

Fig.~\ref{fig:flowchart}.a illustrates an overview of the identified publications. In total, 3740 citations resulted from the bibliographic search in PubMed (n=352), Scopus (n=2798), and Web of Science (n=590) in the Identification phase, from which 935 of the citations were excluded due to duplication. In the screening phase, after carefully reading Tile and the Abstract of the publications, 2692 publications were filtered out because of the topical irrelevance. In the Eligibility phase, 52 publications fulfilled the inclusion and exclusion criteria and ultimately participated in the study (PubMed=27, Scopus=19, and Web of Science=6). Note that half of the eligible publications (n=26) were interestingly published after 2022. Additionally, we included the related review or survey papers, published between 2022 and 2023 (Table~\ref{tab:related_review}). All of the included review papers were articles in peer-reviewed journals.  

\begin{figure}[t]
    \centering
    \includegraphics[width=0.9\linewidth]{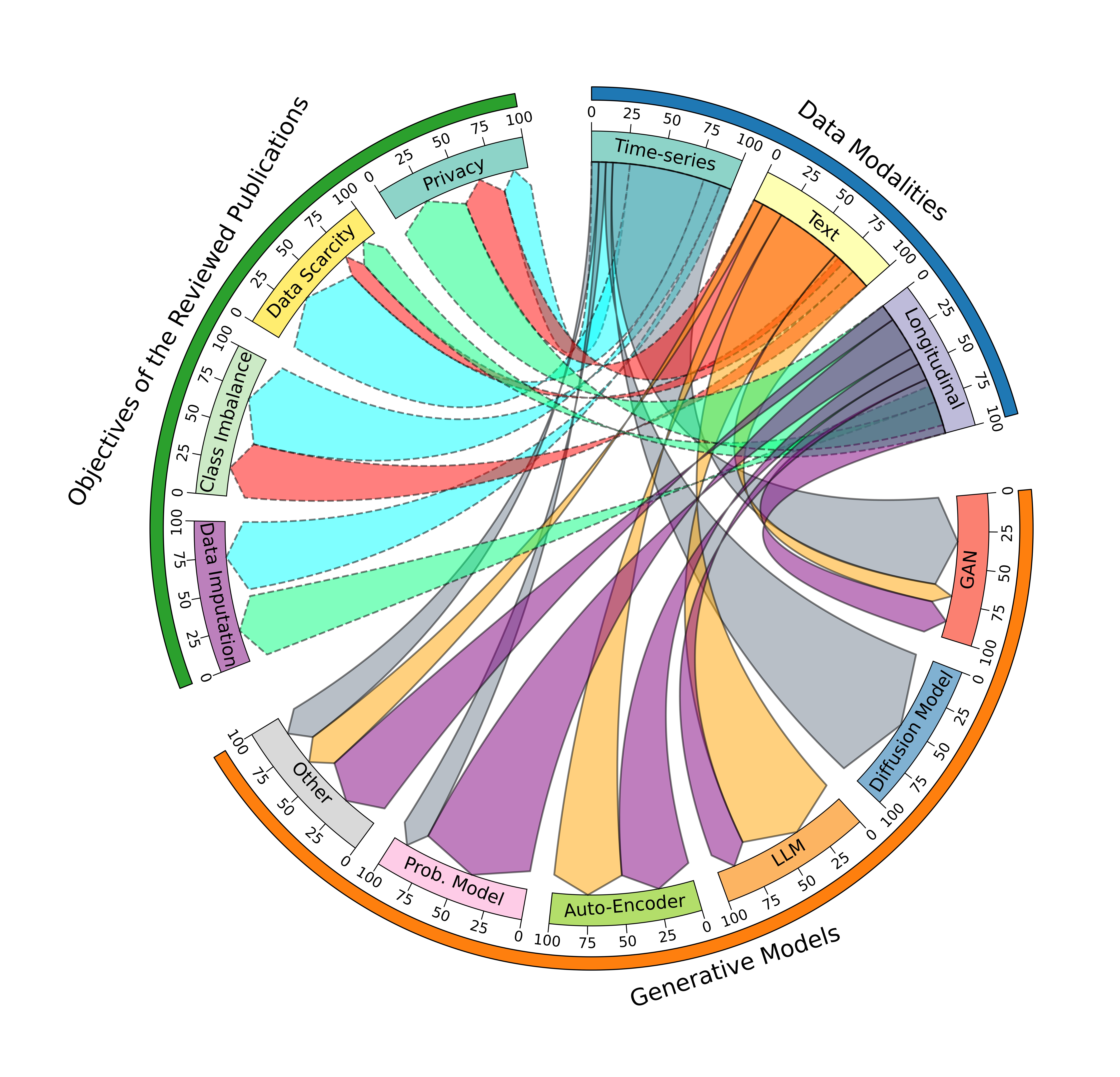}
    \caption{The mutual link between the data modalities, generative models, and research objectives found by the research objectives.}
    \label{fig:overview}
\end{figure}

Fig.~\ref{fig:overview} represents an overview of the findings. It is observed that the GAN-based methods were dominantly employed for generating medical time series, and much less, for generating the longitudinal and the text modalities, respectively. The diffusion model was merely used for this data modality. Although LLMs have been well-received mainly for generating synthetic texts, their application in generating longitudinal data showed promising results. The VAE method was used in a minority of the studies, equally for generating longitudinal and text data, but not so for the time series. The probabilistic models, e.g. Bayesian network, were mostly used for the longitudinal data, and trivially for the time series. 

\subsection{Medical Time Series Data}
\label{src:results:analyses1}

Generation of synthetic physiological time series was observed in $22$ reports ($42\%$) of the published peer-reviewed papers, from which synthetic Electrocardiogram is the most common case study ($10$ studies). Electroencephalogram was the main topic of the second most common objective where the diffusion model resulted in the optimal utility \cite{alcaraz2023diffusion}. Data scarcity and privacy were found to be the main objectives of $12$ and $6$ studies, respectively. Class imbalance and imputation constitute the objective of the rest of the $4$ studies. A great majority of the studies relied on the various fashions of the GAN-based methods where statistical models, such as the diffusion model and hidden Markov model were seen in a minority of $10\%$ of the studies. Table~\ref{tab:timeseries_papers} lists the findings of the survey on the generative models for synthetic time series.

\begin{xltabular}{\textwidth}{b{50pt}|b{50pt}|b{100pt}|b{50pt}|b{130pt}}
\caption{The reviewed publications on generating synthetic medical records of time series.} \label{tab:timeseries_papers} \\

\toprule \textbf{Paper \& Year} & \textbf{Study Objective} &  \textbf{Case Study} &  \textbf{Method} &  \textbf{Key Takeaways}\\ \midrule 
\endfirsthead

\multicolumn{5}{c}%
{\tablename\ \thetable{} -- Continued from the previous page...} \\
\toprule \textbf{Paper \&  Year} & \textbf{Study Objective} &  \textbf{Case Study} &  \textbf{Method} &   \textbf{Key Takeaways} \\ \midrule 
\endhead

\midrule \multicolumn{5}{r}{{Continued on the next page...}} \\ 
\endfoot

\endlastfoot

\cite{wickramaratne2023sleepsim}, 2023 & Data scarcity & Non rapid eye movement sleep EEG & cGAN  & (-) Lack of proper evaluating the fidelity of generated data \\

\cite{boukhennoufa2023novel}, 2023 & Data scarcity  & Post-stroke assessment & Time series siamese GAN & (+) Generating heterogeneous data by avoiding mode collapse \\

\cite{yang2023ts}, 2023 & Data scarcity & ECG arrhythmia, normal \& abnormal ECGs & Time series GAN & (-) Requiring additional experiments to determine the optimal hyperparameters of LSTM \\

\cite{haleem2023deep}, 2023 & Data scarcity & Human physical activities  &  Temporally correlated multi-modal GAN & (+) Preserving temporal correlation between variables learned from multi-modal irregular data  \\

\cite{festag2023medical}, 2023 & Data imputation & Bivariate time series of arterial blood pressure and ECG & Recurrent cGAN & (+) Handling missing time series with different sizes and positions \\

\cite{alcaraz2023diffusion}, 2023 & Privacy & Normal \& abnormal ECGs & Diffusion model & (+) Outperforming GAN-based competitors by capturing long-term dependencies \\

\cite{Asadi2023-mz}, 2023 & Class imbalance & ECG record with paroxysmal atrial fibrillation & GAN & (+) Using physician knowledge to verify generated synthetic samples \\

\cite{li2022tts}, 2022 & Data scarcity & Activity recognition, simulated sinusoidal waves, normal \& abnormal ECGs & Transformer-based time series GAN (TTS-GAN) & (+) Effective handling very long time series \\

\cite{wang2022wearable}, 2022 & Data scarcity & Parkinson disease, human physical activities & Multi-axial cGAN (SensoryGANs) & (+) Synthesizing long multi-axial wearable sensor data \\

\cite{foomani2022synthesizing}, 2022 & Data scarcity & Venous leg ulcer, arterial ulcer, diabetic foot ulcer & Time series conditional Wasserstein GAN & (+) Eliminating the need for access to a big EHR dataset \\

\cite{bing2022conditional}, 2022 & Privacy & ICU records (oxygen saturation, HR, respiratory rate, CO$_2$, etc.) in the MIMIC III database & Time series GAN (HealthGen) & (+) GAN provides better results compared to RNN and Kalman VAE \\

\cite{habiba2021ecg}, 2021 & Privacy & NSR, ECG arrhythmia  & NODE-based GAN & (-) Lack of quantitative evaluations \\

\cite{brophy2021multivariate}, 2021 & Privacy & NSR, ECG arrhythmia & Multivariate GAN & (+) Avoiding GAN mode collapse \\

\cite{lee2021contextual}, 2021 & Data imputation & EEG with sleep stage labels & GAN & (+) Preserving contextual latent in formations \\

\cite{maweu2021generating}, 2021 & Class imbalance, data scarcity & NSR, ECG arrhythmia, EEG data on genetic predisposition to alcoholism & Guided evolutionary synthesizer & (+) Using a non-differentiable objective function\\

\cite{lee2020improved}, 2020 & Data scarcity & ICU records (Oxygen saturation, HR, MAP) & LSTM-based controllable GAN with spectral normalization & (+) Avoiding mode collapse with spectral normalization (-) \\

\cite{li2020activitygan}, 2020 & Data scarcity & Human physical activities (walk, jog, etc.) & GAN & (-) Evaluations are limited to visualization checks \\

\cite{kiyasseh2020plethaugment}, 2020 & Class imbalance & PPG abnormal & cGAN & (+) Evaluating several cGAN model \\ 

\cite{brophy2020synthesis}, 2020 & Privacy & ECG arrhythmia & Multivariate GAN & (+) Using a mini-batch discrimination layer for avoiding mode collapse \\

\cite{wang2020part}, 2020 & Privacy & EEG, ICU records (Oxygen saturation, HR, MAP) & Conditional \& Temporal GAN (PART-GAN)& (+) Integrating differential privacy paradigms with generating noisy and irregularly-sampled time series \\

\cite{nikolaidis2019augmenting}, 2019 & Data scarcity & Obstructive sleep apnea & Concatenating multiple GAN models & (+) Avoiding mode collapse by concatenating results of multiple GANs \\

\cite{dahmen2019synsys}, 2019 & Data scarcity & Smart home-based activity data & Combination of HMM and regression algorithm (SynSys) & (+) Benefiting from HMM's sequence-generative nature \\
\bottomrule
\end{xltabular}

\subsection{Medical Longitudinal Data}
\label{src:results:analyses3}

Table~\ref{tab:longitudinal_papers} lists the survey's findings on the generative models for synthetic longitudinal data. As can be seen, privacy is the main objective of the $16$ out of the $17$ studies with various case studies comprising kidney diseases, patients with hearing loss, Parkinson's and Alzheimer's diseases, chronic heart failure disease, diabetes, hypertension, and hospital admissions. The GAN-based methods were observed in the great majority of the studies as the optimal model outperforming the baseline studies.

\begin{xltabular}{\textwidth}{b{50pt}|b{50pt}|b{100pt}|b{50pt}|b{130pt}}
\caption{The reviewed publications on generating synthetic medical records of longitudinal data.}
\label{tab:longitudinal_papers}  \\

\toprule \textbf{Paper \&  Year} & \textbf{Study Objective} &  \textbf{Case Study} &  \textbf{Method} &   \textbf{Key Takeaways}\\ \midrule 
\endfirsthead

\multicolumn{5}{c}%
{\tablename\ \thetable{} -- Continued from the previous page...} \\
\toprule \textbf{Paper \&  Year} & \textbf{Study Objective} &  \textbf{Case Study} &  \textbf{Method} &  \textbf{Key Takeaways} \\ \midrule 
\endhead

\midrule \multicolumn{5}{r}{{Continued on the next page...}} \\ 
\endfoot

\endlastfoot

\cite{theodorou2023synthesize}, 2023 & Privacy, data scarcity &  Hospital visits from MIMIC III database & Hierarchical auto-regressive language model & (+) Fidelity of SHR is improved by utilizing a probabilistic and an autoregressive model for estimating longitudinal data at the visit and code level \\

\cite{li2023improving}, 2023 & Data scarcity, privacy & multi-dimensional cancer and type-2 diabetes data & GAN-boosted semi-supervised learning & (+) Utilizes the underlying graphical structure of EHRs \\

\cite{li2023generating}, 2023 & Privacy, data scarcity &  EHR time series for ICU patients & Mixed-type longitudinal GAN & (+) Generating mixed-type time series by effectively capturing the temporal characteristics of the original data \\

\cite{nikolentzos2023synthetic}, 2023 & Privacy & Critical care patients data admitted to ICU (e.g. \#visits, diagnosis, etc) from MIMIC IV dataset & Variational graph auto-encoder & (+) Generating synthetic patient trajectories from EHRs with graph learning \\

\cite{mosquera2023method}, 2023 & Privacy & Longitudinal health records (e.g. age, vital statistics) & RNN & (-) Generating lengthy sequences has limitations \\

\cite{wang2022using}, 2022 & Data scarcity & Type-2 diabetes data & Generative Markov-Bayesian-based model & (-) limited to a single chronic disease and using only ICD-10 data code \\

\cite{shi2022generating}, 2022 & Privacy & Health records of patients with hypertension & GAN & (-) The criteria for data inclusion and exclusion could potentially result in selection bias \\

\cite{wendland2022generation}, 2022 & Privacy, data imputation & Parkinson’s disease and Alzheimer’s disease & Multi-modal Neural Ordinary Differential Equations & (+) Handling multi-modal data along with learning continuous-time real data trajectories (-) Limited to the static categorical variables \\

\cite{wang2022promptehr}, 2022 & Privacy & Hospital visits from MIMIC III database & GPT-2 & (+) Formulating the generation of the heterogeneous EHRs as a text-to-text translation task using LLMs \\

\cite{zhou2022datasifter}, 2022 & Privacy, data imputation & Hospital visits from MIMIC III database &  DataSifter-II (ruled-based method) & (+) Improved privacy of the time-varying correlated data by using a generalized linear mixed model and random effects-expectation maximization tree  \\

\cite{kaur2021application}, 2021 & Privacy & Hospital visits from MIMIC III database & Bayesian network (-) & Struggling to preserve multivariate relationships in the datasets \\

\cite{abell2021generating}, 2021 & Privacy & Acute kidney injury & GAN & (-) Insufficient evaluation of the fidelity and the utility\\

\cite{zhang2021synteg}, 2021 & Privacy & The EHR from type-2 diabetes, heart failure, and hypertension & GAN & (+) Mitigation of the GAN issues by using a two-step learning method: dependency learning and conditional simulation \\

\cite{lee2020generating}, 2020 & Privacy & Hospital visits from MIMIC III database & Adversarial auto-encoder & (+) Adversarially learning both the continuous latent distribution and the discrete data distribution \\

\cite{yoon2020anonymization}, 2020 & Privacy & Chronic heart failure, organ transplantation & cGAN & (+) Improved privacy; the identifiability of the SHR is quantified and employed for the optimization of a cGAN \\

\cite{christensen2019fully_body}, 2019 & Privacy & Hearing loss patients & Bayesian network & (-) Insufficient evaluation of the fidelity and the utility \\

\cite{baowaly2019synthesizing}, 2019 & Privacy & Hospital visits from MIMIC III database & GAN & (-) Limited to generating discrete synthetic EHRs \\
\bottomrule
\end{xltabular}

\subsection{Medical Text Data}
\label{src:results:analyses2}

Medical texts are an important part of an EHR, reflecting the medical assessments and decisions. Reading these texts can put the underlying EHR at risk of identification, and therefore, privacy is regarded as an important objective when it comes to the SHR. This was confirmed by our study as privacy was the objective of the $9$ studies out of the $12$ studies that participated in this survey with various case studies. Table~3 lists the findings of the survey on the generative models for synthetic text data.

In contrast with the synthetic health time series generation where the proposed models were dominantly based on the GAN, GPT-style models are employed in approximately 40\% of studies, surpassing individual usage rates of other generative methods. This is justifiable considering the versatility of GPT models. Such elaborative capabilities of GPT provided the ability to generate synthetic medical texts in different languages, spanning from far eastern countries, e.g. Chinese, to European countries, e.g. Dutch, and English (see the case studies in Table~\ref{tab:text_papers}). Nevertheless, the GAN-based models were seen in some studies.

\begin{xltabular}{\textwidth}{b{50pt}|b{50pt}|b{100pt}|b{50pt}|b{130pt}}
\caption{The reviewed publications on generating synthetic medical records of text data.}
\label{tab:text_papers}  \\ \toprule

\textbf{Paper \&  Year} & \textbf{Study Objective} &  \textbf{Case Study \& Language} &  \textbf{Method} & \textbf{Key Takeaways}\\ \midrule 
\endfirsthead

\multicolumn{5}{c}%
{\tablename\ \thetable{} -- Continued from the previous page...} \\
\toprule \textbf{Paper \&  Year} & \textbf{Study Objective} &  \textbf{Case Study \& Language} &  \textbf{Method} & \textbf{Key Takeaways} \\ \midrule 
\endhead

\midrule \multicolumn{5}{r}{{Continued on the next page...}} \\ 
\endfoot

\endlastfoot

\cite{peng2023study}, 2023 & Privacy & English clinical notes from inpatient, outpatient, and emergency settings & GPT-3 & (+) The physicians’ Turing test shows readability and clinical relevance of generated clinical text \\

\cite{hiebel2023can}, 2023 & Privacy & clinical case corpus in French (e.g. medical history of patients, treatment received at the hospital) & GPT-2, BLOOM & (-) Lack of the privacy leakage evaluation \\

\cite{khademi2023data}, 2023 & Data scarcity & Emergency department notes in English (e.g. medical history, nurse’s observations, etc) & GPT-2 & (+) GPT-2 outperformed word swap and word-embedding techniques \\

\cite{wang2023differentially}, 2023 & Privacy, security & Medical description data in English from CMS\footnote{\url{https://www.cms.gov/}} public healthcare records & Recurrent VAE & (+) Extending synthetic text data generator to the federated learning
scenario \\

\cite{zhou2022datasiftertext}, 2022 & Privacy & Work-related injury description records in English, hospital visits from MIMIC III database & Perturbation-based data sifting & (+) Generating synthetic datasets with individual-level data obfuscation while maintaining population-level information  \\

\cite{shim2021synthetic}, 2021 & Data scarcity & English texts with temporal information of sleep-related activities & BERT & (+) Extracting temporal information from the user-generated text \\

\cite{kasthurirathne2021generative}, 2021 & Privacy & Laboratory messages in English pertaining to Salmonella & Sequence GAN & (-) Ineffective generating long laboratory messages \\ 

\cite{al2021differentially}, 2021 & Privacy & Hospital reports in English from MIMIC III database &  GPT-2 & (+) Generating synthetic texts with the removed personal information and differential privacy \\

\cite{libbi2021generating}, 2021 & Privacy & Medical Dutch text reports from elderly care, mental care, and disabled care domains & LSTM, GPT-2 & (+) LSTM produces synthetic text with higher precision compared to GPT-2, on the other hand, GPT-2 generates more coherent samples \\

\cite{guan2019method}, 2019 & Privacy, Class imbalance & Chinese EHR texts (e.g. personal information, history of present and past illnesses, etc) & cGAN & - \\

\cite{kasthurirathne2019adversorial}, 2019 & Privacy & Text cancer pathology reports in English & Sequence GAN & (-) Lack the privacy leakage evaluation )\\

\cite{syed2019gender}, 2019 & Class imbalance & Gender prediction based on user clicks on articles of a health-based website & Sequence GAN & (+) Sequence GAN outperforms the minority oversampling technique\\
 
\cite{lee2018natural}, 2018 & Privacy & $\approx$5.8M visit records (e.g. age, gender, discharge diagnosis code) in English & Encoder-Decoder LSTM & - \\
\bottomrule
\end{xltabular}

\section{Discussion}
\label{src:discussion}

\subsection{Clinical Applicability}
\label{src:discussion:clinical_applicability}

The need for a rich dataset for training ML methods on the one hand, and the difficulties in collecting patient data, e.g. privacy issues, on the other hand, make the generation of SHR a practical strategy. This is subjected to a high level of security in terms of re-identification along with acceptable fidelity. Countries adopt different regulations that intensively restrict the sharing of patient data, which is sometimes administered in a federated way. The use of SHR allows sharing of data, which can be employed by researchers to develop advanced ML methods for different medical applications, where access to the real data is problematic. Another application of SHR is the cases in which a heavy class imbalance negatively affects the learning process. In such cases, generative models are employed to create synthetic medical data for the minority classes. This is different from data augmentation where the minority data is reproduced, as the statistical distribution of the data is taken into account for generating SHRs.

\subsection{Distinction to the Other Review Studies}
\label{src:discussion:review_papers}

Several surveys and review studies have been previously conducted on different models for generating SHRs. However, a comprehensive study on this topic with sufficient pervasiveness to explore different aspects of the studies, cannot be found in the existing literature, from a practical perspective. In addition, unlike other review papers (Table~\ref{tab:related_review}), ours covers more data modalities and DL models, thereby providing readers with novel perspectives on the topic. The outcomes of such a pervasive study can unveil practical limitations and bottlenecks of the existing methodologies to choose an appropriate model for such a demanding application.

\begin{table}[htbp]
\centering
\caption{Review publications on synthetic health data generation.}
\label{tab:related_review} 
\resizebox{\textwidth}{!}{
\begin{tabular}{cccccccccccccc}
\toprule
\multirow{3}{*}{\textbf{Paper}} &  \multirow{3}{*}{\textbf{Year}} &  \multicolumn{3}{c|}{\textbf{Data Modality}} 
&  \multicolumn{5}{c|}{\textbf{DL Model}} & \multicolumn{4}{c}{\textbf{Study Objective}}\\ \cmidrule{3-14}
&  & \textbf{Time-} & \multirow{2}{*}{\textbf{Text}} & \multicolumn{1}{c|}{\multirow{2}{*}{\textbf{Longitudinal}}} & \multirow{2}{*}{\textbf{GAN}} & \multirow{2}{*}{\textbf{LLM}} & \multirow{2}{*}{\textbf{Diff.}} & \textbf{Auto-} & \textbf{Prob. Gen.} & \multicolumn{1}{|c}{\textbf{Data}} & \multirow{2}{*}{\textbf{Privacy}} & \textbf{Class} & \textbf{Data}  \\
  
&  & \multicolumn{1}{c}{\textbf{Series}} & \textbf{} & \multicolumn{1}{c|}{\textbf{}} & \textbf{} & \textbf{} & \textbf{} & \textbf{Encoder} & \textbf{Model} & \multicolumn{1}{|c}{\textbf{Imputation}} & \textbf{} & \textbf{Imbalance} & \textbf{Scarcity} \\ \midrule

\cite{murtaza2023synthetic} & 2023 & \greencheck & \greencheck & \greencheck & \greencheck & \crossmark & \crossmark & \greencheck & \crossmark & \crossmark & \greencheck & \crossmark & \crossmark \\
\cite{koo2023survey} & 2023 & \greencheck & \crossmark & \greencheck & \crossmark & \crossmark & \greencheck & \crossmark & \crossmark & \greencheck & \crossmark & \crossmark & \greencheck  \\
\cite{hernandez2022synthetic} & 2022 & \greencheck & \crossmark & \crossmark & \greencheck & \crossmark & \crossmark & \greencheck & \crossmark & \crossmark & \greencheck & \crossmark & \greencheck \\
\cite{li2022neural} & 2022 & \crossmark & \greencheck & \crossmark & \greencheck & \greencheck & \crossmark & \greencheck & \crossmark & \crossmark & \greencheck & \crossmark & \greencheck \\
\cite{hahn2022contribution} & 2022 & \crossmark & \crossmark & \crossmark & \greencheck & \crossmark & \crossmark & \greencheck & \crossmark & \crossmark & \greencheck & \greencheck & \greencheck \\
\cite{ghosheh2024survey} & 2022 & \greencheck & \crossmark & \greencheck & \greencheck & \crossmark & \crossmark & \crossmark & \crossmark & \greencheck & \greencheck & \greencheck & \greencheck \\
\midrule
Ours & 2023 & \greencheck & \greencheck & \greencheck & \greencheck & \greencheck & \greencheck & \greencheck & \greencheck & \greencheck & \greencheck & \greencheck & \greencheck \\ \bottomrule
\end{tabular}
}
\end{table}

\subsection{Limitations of ML Methods in Generating Synthetic Data}
\label{src:discussion:DL}

This study provided a scoping review of the most popular generative models for producing SHRs. Our analysis showed that the researchers employed GAN-based models most for generating synthetic time series compared to the other alternatives (See Supplementary Note~1). In addition to the GANs, the probabilistic models were widely used for generating synthetic longitudinal data. However, several studies reported that GAN-based models generally suffered from \begin{mylist}
\item the mode collapse issue \cite{boukhennoufa2023novel, brophy2021multivariate, brophy2020synthesis, lee2020improved, nikolaidis2019augmenting},
\item requiring preliminary experiments to identify optimal hyperparameters, and 
\item having biases towards high-density classes \cite{alcaraz2023diffusion}.
\end{mylist} Diffusion models demonstrated promising results in synthesizing time series compared to GANs. Nevertheless, resolving expensive computational costs and interoperability difficulties of diffusion models, are considered ongoing research endeavors. Finally, current works could gain significant advantages by integrating domain-specific expertise from physicians into the learning process \cite{Asadi2023-mz, koo2023survey}.

Generating synthetic clinical notes is a less explored area in the literature. Recent advancements of LLMs \cite{khademi2023data, hiebel2023can, peng2023study} have demonstrated significant improvements in generating synthetic clinical notes. Regardless, LLMs suffer from requiring massive processing power (\cite{peng2023study} leveraged 560 A100 GPUs for 20 days to train the LLM). Furthermore, \cite{huang2022towards, valmeekam2022large} addressed that LLMs struggle with complex reasoning problems. Non-reasoned outputs for generating synthetic clinical notes lacked coherence, consistency, and certainty \cite{huang2022towards}. Chain-of-Thought prompting \cite{wei2022chain} standed out as a leading method aimed to improve complex reasoning capabilities through intermediate suggestions. While Chain-of-Thought has shown promising results, its effectiveness in tackling the reasonability of LLMs for complex multi-modal input and tasks necessitating compositional generalization remains an unresolved problem \cite{huang2022towards}. Despite the success of generative models, tweaking effort was sometimes required to achieve optimal performance \cite{tornede2024automl}. It is worth mentioning that despite the success of studied papers in generating SHRs, the reproducibility of results of several studies is under question as 
\begin{mylist}
    \item the implementation code is not available, and 
    \item details of training hyper-parameters have not been reported.
\end{mylist}  

\subsection{Available Training Datasets}
\label{src:discussion:dataset}

Generating SHR necessitates real datasets to train the generative model, and the quality of the training data accessible defines the caliber of synthetic data \cite{bandi2023power, abbasian2024foundation, alaa2022faithful, ghosheh2024survey, el2022utility, kaabachi2022generation}. The EHRs collected at healthcare sites are usually multi-dimensional and longitudinal datasets, recording patient history over multiple visits. However, secondary use of this data is restricted by privacy laws \cite{hernandez2022synthetic, jordon2021hide, van2023beyond}; nevertheless, there are several de-identified datasets available for generating synthetic data. The popular databases for generating SHRs used by eligible publications of this review study can be found in Table~\ref{tab:Datasets}. Our findings show that despite the existence of public datasets available for generating SHRs, the majority of public longitudinal medical datasets primarily focus on ICU records, prioritizing acute patient cases and overlooking non-acute medical conditions. Furthermore, these public datasets often lack a comprehensive representation of all demographic groups and geographic regions, which limits their relevance and generalizability to broader populations. Finally, it is worth noting that most longitudinal records are reported in English, underscoring the current shortage of public resources across diverse languages.

\begin{xltabular}{\textwidth}{b{50pt}b{20pt}b{140pt}b{30pt}b{20pt}b{40pt}b{50pt}}
\caption{Summary of public databases for generating SHRs. }
\label{tab:Datasets}
\\ \toprule
\multirow{3}{*}{\textbf{Dataset}} & \multirow{3}{*}{\textbf{URL}} & \multirow{3}{*}{\textbf{Case Study}} & \multicolumn{3}{c}{\textbf{Data Modality}} & \multirow{3}{*}{\textbf{Ref. Study}} \\ \cmidrule{4-6} 

&  & & \textbf{Time Series} &  \textbf{Text} & \textbf{Longitud.} & \\ \midrule   
\endfirsthead

\multicolumn{7}{c}
{{ \tablename \thetable{} -- Continued from previous page...}} \\
 \toprule

\multirow{3}{*}{\textbf{Dataset}} & \multirow{3}{*}{\textbf{URL}} & \multirow{3}{*}{\textbf{Case Study}} &  \multicolumn{3}{c}{\textbf{Data Modality}} & \multirow{3}{*}{\textbf{Ref. Study}}\\ \cmidrule{4-6} 

&   &   & \textbf{Time Series} &  \textbf{Text} & \textbf{Longitud.} & \\ \midrule 
\endhead

\multicolumn{7}{r}{{Continued on the next page...}} \\ 
\endfoot

\endlastfoot

MIMIC III \cite{johnson2016mimic} & \href{https://physionet.org/content/mimiciii/1.4/}{Link} & De-identified patient records admitted to intensive care unit (ICU) from 2001 to 2012 containing demographics, laboratory test results, procedures, caregiver notes, medications, imaging reports, mortality, etc. & \multicolumn{1}{c}{\greencheck} & \multicolumn{1}{c}{\greencheck} & \multicolumn{1}{c}{\greencheck} & \cite{ bing2022conditional,  theodorou2023synthesize,  wang2022promptehr,  zhou2022datasifter,  kaur2021application,  lee2020generating,  baowaly2019synthesizing,  zhou2022datasiftertext,  al2021differentially,  li2023generating} \\ \midrule 

MIMIC IV \cite{johnson2023mimic} & \href{https://physionet.org/content/mimiciv/2.2/}{Link} & Critical care data for patients admitted to the ICU at the Beth Israel Deaconess Medical Center. The information available includes patient measurements, orders, diagnoses, procedures, treatments, and clinical notes  & \multicolumn{1}{c}{\greencheck} &  \multicolumn{1}{c}{\greencheck} & \multicolumn{1}{c}{\greencheck}  & \cite{ nikolentzos2023synthetic} \\ \midrule

eICU \cite{pollard2018eicu} & \href{https://physionet.org/content/eicu-crd/2.0/}{Link} & A critical care database spanning multiple centers holds information from over 200,000 admissions to ICUs from 208 hospitals situated across the U.S. & \multicolumn{1}{c}{\greencheck} & \multicolumn{1}{c}{\greencheck} & \multicolumn{1}{c}{\greencheck} &  \cite{ lee2020improved,  wang2020part,  li2023generating} \\ \midrule 

HiRID \cite{yeche2021hirid} & \href{https://physionet.org/content/hirid/1.1.1/}{Link} & a high-resolution ICU dataset relating to more than 3 billion observations from $\approx$34,000 ICU patient admissions & \multicolumn{1}{c}{\greencheck} & & \multicolumn{1}{c}{\greencheck} &  \cite{ li2023generating} \\ \midrule

Pile \cite{gao2020pile} & \href{https://pile.eleuther.ai/}{Link} & An 825 GiB English text corpus from 22 diverse high-quality subsets including the medical domain & & \multicolumn{1}{c}{\greencheck} & & \cite{ peng2023study} \\ \midrule 

E3C \cite{zanoli2023assessment} & \href{https://github.com/hltfbk/E3C-Corpus}{Link} & A multilingual (English, French, Italian, Spanish, and Basque) corpus database containing biomedical documents extracted from different sources, including journals, existing biomedical corpora, etc. & & \multicolumn{1}{c}{\greencheck} & & \cite{ hiebel2023can} \\ \midrule 

INPCR \cite{mcdonald2005indiana} & \href{https://www.regenstrief.org/rds/data/}{Link} & This database stands as one of the most extensive health information exchange in U.S., encompassing more than 100 distinct healthcare organizations contributing data. The database contains information on over 18 million patients, comprising 10 billion clinical observations, and over 147 million text reports. & & \multicolumn{1}{c}{\greencheck} & & \cite{ kasthurirathne2021generative,  kasthurirathne2019adversorial} \\ \midrule 

Administrative Health Records \cite{sharma2019characterisation} & \href{https://open.alberta.ca/dataset/overview-of-administrative-health-datasets}{Link} & Patients received a prescription for an opioid during the 7-year study window. Data includes demographic information, laboratory tests, prescription history, hospitalizations, etc. &  &  & \multicolumn{1}{c}{\greencheck} & \cite{ mosquera2023method} \\ \midrule

PPMI \cite{marek2011parkinson} & \href{https://www.ppmi-info.org/}{Link} & Observational clinical study containing 354 Parkinson patients who participated in a range of clinical, neurological, and demographic assessments  &  &  & \multicolumn{1}{c}{\greencheck}  &  \cite{ wendland2022generation} \\ \midrule

NACC \cite{beekly2007national} & \href{https://naccdata.org/}{Link} & Storing patient-level Alzheimer’s disease data collected from 2284 patients across multiple clinics  &  &  &  \multicolumn{1}{c}{\greencheck} & \cite{ wendland2022generation} \\ \midrule

Evotion \cite{christensen2019fully} & \href{https://zenodo.org/records/2668210}{Link} & Information relating to patterns of real-world hearing aid usage and sound environment exposure. EVOTION contains longitudinally observations from 53 individuals and includes the following measures: the sound environment, the hearing aid setting, timestamps, ID, and the degree of hearing loss on the best hearing ear of the individuals &  &  & \multicolumn{1}{c}{\greencheck} &  \cite{ christensen2019fully_body} \\ \midrule

SEER \cite{hankey1999surveillance} & \href{https://seer.cancer.gov/}{Link} & This database provides information on cancer incidence and survival from population-based cancer registries in 22 U.S. geographic areas &  &  & \multicolumn{1}{c}{\greencheck} &  \cite{ li2023improving} \\ \midrule

Human Activity Sensing Archive \cite{kawaguchi2011hasc} & \href{http://hasc.jp/}{Link} & A large-scale human activity corpus archive &  \multicolumn{1}{c}{\greencheck} &  &  & \cite{ wang2022wearable,  li2020activitygan} \\ \midrule

UCR Time Series Archive \cite{dau2019ucr} & \href{https://www.cs.ucr.edu/~eamonn/time_series_data_2018/}{Link} & A large repository of time series datasets including health records of Electrocardiogram (ECG), motion, etc. &  \multicolumn{1}{c}{\greencheck} &  &  & \cite{ yang2023ts} \\ \midrule

Autonomic Aging \cite{schumann2022autonomic} & \href{https://physionet.org/content/autonomic-aging-cardiovascular/1.0.0/}{Link} & A database of high-resolution biological signals to describe the effect of healthy aging on cardiovascular regulation & \multicolumn{1}{c}{\greencheck} &  &  & \cite{ festag2023medical} \\ \midrule

PTB-XL \cite{wagner2020ptb} & \href{https://physionet.org/content/ptb-xl/1.0.3/}{Link} & A large dataset of 21799 clinical 12-lead ECGs from 18869 patients. The raw waveform data was annotated by up to two cardiologists &  \multicolumn{1}{c}{\greencheck} &  &  & \cite{ alcaraz2023diffusion} \\ \midrule

AF Classification Challenge \cite{clifford2017af} & \href{https://physionet.org/content/challenge-2017/1.0.0/}{Link} & A dataset of single-lead ECGs (between 30 s and 60 s in length) with normal sinus rhythm, atrial fibrillation (AF), an alternative rhythm, or noisy classes &  \multicolumn{1}{c}{\greencheck} &  & & \cite{ Asadi2023-mz} \\ \midrule

UniMiB \cite{micucci2017unimib} & \href{http://www.sal.disco.unimib.it/technologies/unimib-shar/}{Link} & A dataset tailored for human activity recognition and fall detection with 11,771 acceleration samples performed by 30 subjects aged between 18 and 60 years & \multicolumn{1}{c}{\greencheck} &  &  & \cite{ li2022tts} \\ \midrule

PAMAP2 \cite{reiss2012introducing} & \href{https://archive.ics.uci.edu/dataset/231/pamap2+physical+activity+monitoring}{Link} & The PAMAP2 Physical Activity Monitoring dataset contains data of 18 different physical activities (e.g. walking, cycling, playing soccer), performed by 9 subjects wearing 3 inertial measurement units and a heart rate monitor & \multicolumn{1}{c}{\greencheck} &  &  & \cite{ wang2022wearable} \\ \midrule

MIT-BIH Arrhythmia \cite{goldberger2000physiobank} & \href{https://physionet.org/content/mitdb/1.0.0/}{Link} & This dataset contains 48 30-min excerpts of two-channel ECG recordings, obtained from 47 subjects studied between 1975 and 1979  &  \multicolumn{1}{c}{\greencheck} &  &  & \cite{ habiba2021ecg,  maweu2021generating,  brophy2020synthesis} \\ \midrule

MIT-BIH Normal Sinus Rhythm (NSR) \cite{goldberger2000physiobank} & \href{https://www.physionet.org/content/nsrdb/1.0.0/}{Link} & This dataset includes 18 long-term ECG recordings of subjects with no significant arrhythmias &  \multicolumn{1}{c}{\greencheck} &  &  & \cite{ habiba2021ecg} \\ \midrule

Sleep-EDF (Expanded) \cite{kemp2000analysis} & \href{https://www.physionet.org/content/sleep-edf/1.0.0/}{Link} & This dataset contains 197 whole-night PolySomnoGraphic sleep recordings, containing EEG, Electrooculography, chin electromyography, and event markers. Some records also contain respiration and body temperature & \multicolumn{1}{c}{\greencheck} &  &  & \cite{ lee2021contextual} \\ \midrule

The National Sleep Research Resource \cite{zhang2018national} & \href{https://sleepdata.org/}{Link} & Compilation of annotated sleep datasets, along with interfaces and tools for accessing and analyzing this data, is available & \multicolumn{1}{c}{\greencheck} &  &  & \cite{ wickramaratne2023sleepsim} \\ \midrule

UCI EEG Dataset & \href{https://archive.ics.uci.edu/dataset/121/eeg+database}{Link} & This dataset examines EEG correlates of genetic predisposition to alcoholism. It contains measurements from 64 electrodes placed on the scalp sampled at 256 Hz &  \multicolumn{1}{c}{\greencheck} &  &  & \cite{ maweu2021generating} \\ \midrule

PhysioNet Challenge 2015 \cite{clifford2015physionet} & \href{https://physionet.org/content/challenge-2015/1.0.0/}{Link} & This database includes ECG and Photoplethysmogram (PPG) recordings from 750 patients that suffered either of the following cardiac conditions; asystole, extreme bradycardia, extreme tachycardia, ventricular tachycardia, and ventricular flutter  &  \multicolumn{1}{c}{\greencheck} &  &  & \cite{ kiyasseh2020plethaugment} \\ \midrule

PPG-DB \cite{liang2018new} & \href{https://figshare.com/articles/dataset/PPG-BP_Database_zip/5459299}{Link} & A collection of de-identified photoplethysmography data for studying cardiovascular disease. The dataset contains 657 records from 219 subjects, spanning ages 20 to 89 years, with records of diseases like hypertension and diabetes &  \multicolumn{1}{c}{\greencheck} &  &  & \cite{ kiyasseh2020plethaugment} \\ \midrule

UCI ML Repository \cite{asuncion2007uci} & \href{https://archive.ics.uci.edu/}{Link} & A collection of databases in various subjects including health and medicine & \multicolumn{1}{c}{\greencheck} & \multicolumn{1}{c}{\greencheck} & \multicolumn{1}{c}{\greencheck} & \cite{ boukhennoufa2023novel,  yang2023ts,  li2023improving} \\ \midrule

PhysioNet \cite{goldberger2000physiobank} & \href{https://physionet.org/}{Link} & An extensive collection of health data from both healthy individuals and those dealing with conditions like sudden cardiac death, congestive heart failure, epilepsy, gait disorders, sleep apnea, and aging & \multicolumn{1}{c}{\greencheck} & \multicolumn{1}{c}{\greencheck} & \multicolumn{1}{c}{\greencheck}  & \cite{ li2022tts,  nikolaidis2019augmenting} \\ \bottomrule

\end{xltabular}

\subsection{Challenges and Research Gaps in Evaluating SHRs}
\label{src:discussion:measurements}

One of the objectives of this survey is to help researchers find an optimal generative model for SHR among a great variety of existing ones, which in turn demands a set of objective performance measures for comparison. Various statistical and intuitive metrics have been employed for comparing the performance of generative models which makes choosing an optimal model for a case study complicated. In addition, some of the metrics are based on the discrimination power of the physicians \cite{wickramaratne2023sleepsim, alcaraz2023diffusion, Asadi2023-mz, wang2022using}, whereas some others rely on the performance of a benchmark binary classifier to distinguish between SHR and EHR \cite{lee2020generating, kaur2021application}. 

\begin{figure}[htbp]
    \centering
    \includegraphics[width=0.65\columnwidth]{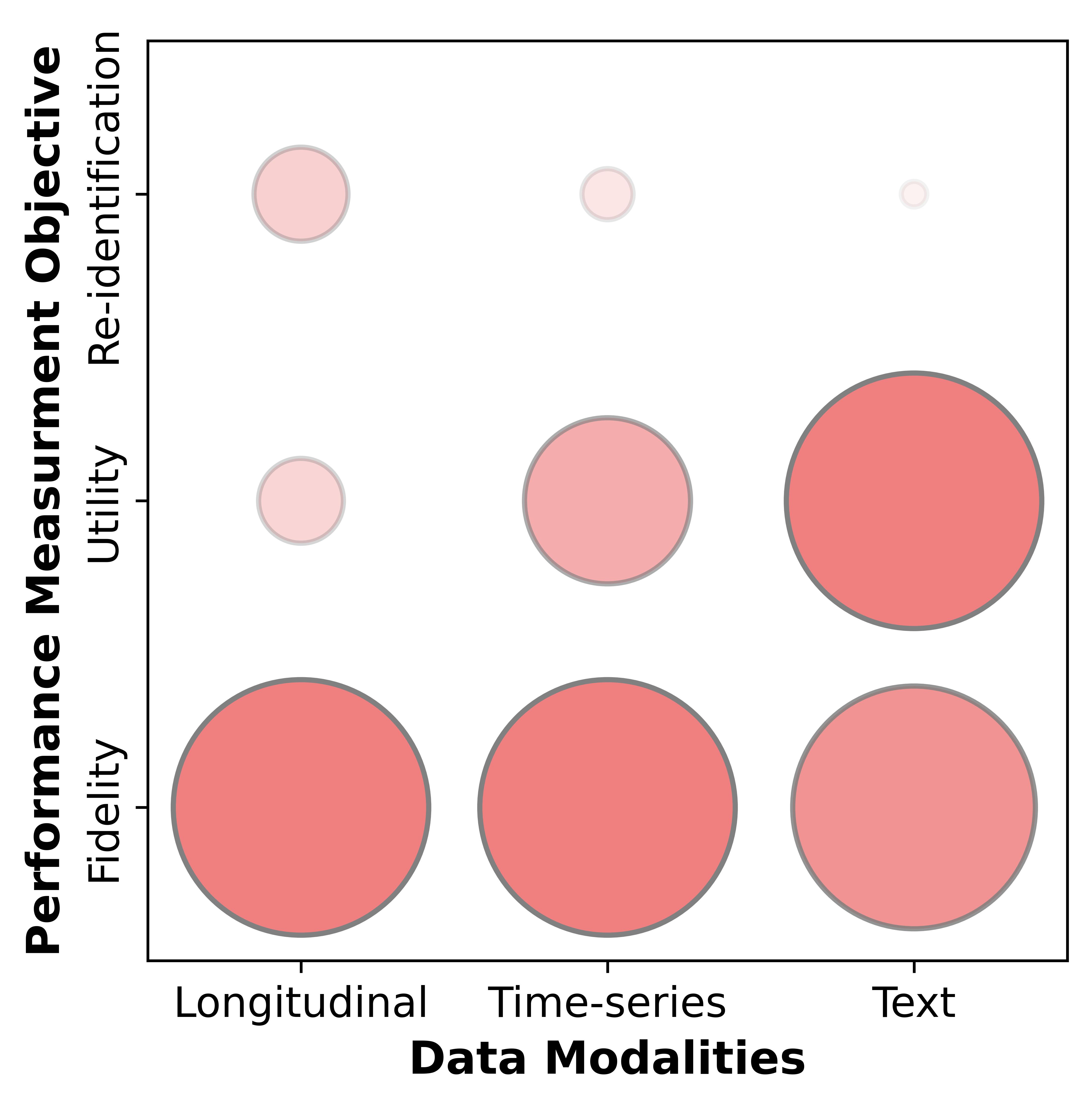}
    \caption{Normalized distribution of performance measurement objectives over the data modalities. Larger circles display more publications in each category. The Figure indicates an inadequate evaluation of re-identification of SHRs in studied papers. In addition, evaluating the utility of longitudinal data has been less researched compared to medical time series and text data.}
    \label{fig:performance}
\end{figure}

An important challenge, observed in this study is the lack of generic methods and metrics for comparing the performance of different generative models. Fig.~\ref{fig:performance} represents the distribution of the three performance measurement objectives: fidelity, utility, and re-identification, over the data modalities. As can be seen, generating high-fidelity SHRs appears to be the main objective for all data modalities. It is implied from Fig.~\ref{fig:performance} that introducing appropriate performance measures to address the privacy of SHR is a contemporary research gap due to the shortage of pertinent studies. This is the case for the utility when it comes to the longitudinal SHR. 

Statistical diversity of the data has been recently introduced as a measure for comparing the utility and fidelity of SHRs \cite{bahrpeyma2021methodology, alaa2022faithful}. Furthermore, fairness of SHR was also defined as another comparative objective for the performance measures \cite{bhanot2021problem}. These two measures were not widely accepted by the researchers according to the citation records. All in all, there are no best established systematic criteria or practice on how to evaluate SHRs.

\subsection{Future Research Trend in GenAI for SHRs}
\label{src:discussion:future_trend}

Generative models for time series prediction showed promising results in various medical applications for classification and identification problems \cite{gharehbaghi2017deep, gharehbaghi2019artificial}. This topic was indeed an initiation for creating SHR, as was previously reported by reviewed studies. Recent scientific endeavors revealed that the application of SHR is not limited to the privacy of patient data, but can be extended towards statistical planning for clinical trials \cite{ghosh2021propensity}, and evermore, towards addressing ethical issues \cite{ive2022leveraging} by solving bias (e.g. black patients were less likely to be admitted to cardiology for heart failure care \cite{eberly2019identification}) in the original dataset. In terms of methodology, the recent methodological trend of the practical models for the creation of SHR shows a shift from the GAN-based methods to the statistical models such as graph neural networks and diffusion models \cite{nikolentzos2023synthetic, alcaraz2023diffusion}, and data fairness was addressed by one of the recent studies \cite{bhanot2021problem}. 

For further reading, we recommend the following key papers that complement our work and provide a deeper understanding of the subject of generating SHR. \cite{abbasian2024foundation} demonstrated that the evaluation metrics currently available for generic LLMs lack an understanding of medical and health-related concepts, which aligns closely with the findings of our study. In addition, the authors introduced a comprehensive collection of LLM-based metrics tailored for the evaluation of healthcare chat-bots from an end-user perspective. Social determinants of health (SDoH) encompass the conditions of individuals' lives, influenced by the distribution of resources and power at various levels, and are estimated to contribute to 80–90\% of modifiable factors affecting health outcomes \cite{guevara2024large}. However, documentation of SDoH is often incomplete in the structured data of EHRs. In \cite{guevara2024large}, the authors extracted six categories of SDoH from the EHR using LLMs to support research and clinical care. To address class imbalance and fine-tune the extraction model, the authors leveraged an LLM to generate synthetic SDoH data.

\section{Key Takeaways}
\label{src:takeaways}

\begin{enumerate}[noitemsep]
    \item Different varieties of the GAN-based models have been proposed for generating synthetic medical time series, and longitudinal data. This puts GAN as the dominant model for these two data modalities, where fidelity is the main objective of the performance measurement for both cases. Nevertheless, the diffusion model outperformed the GAN-based ones in a case study on generating synthetic ECG.
    \item Among the various ML models, LLM received the most popularity for generating synthetic medical text, where the utility was found to be the major performance measurement explored as the study objective.
    \item Privacy was observed to be the dominant objective of creating SHR, even though other objectives such as class imbalance and data scarcity were widely studied. This is in line with the directives of the European Union on releasing the first regulation of artificial intelligence. 
    \item A broad range of metrics was employed as the performance measures for different study objectives, making the methodological comparison questionable. It seems the development of the appropriate evaluation metrics is considered as a major research gap.
\end{enumerate}

\appendix

\section{Abbreviation}
\label{src:appendix:acronyms}

Table~\ref{tab:acronym} lists acronyms in this study.

\begin{xltabular}{\textwidth}{b{80pt}b{230pt}}
\caption{Alphabetic Sorting of Acronyms.}
\label{tab:acronym} 

\\ \toprule

\textbf{Acronym} & \textbf{Description}  \\ \midrule 
\endfirsthead

\multicolumn{2}{c}
{{ \tablename \thetable{} -- Continued from previous page}} \\
\midrule 

\textbf{Acronym} & \textbf{Description}  \\ \midrule 
\endhead

\midrule \multicolumn{2}{r}{{Continued on next page}} \\ 
\endfoot

\endlastfoot
AIA & Attribute Inference Attack \\

AUPRC & Area Under the Precision-Recall Curve \\

AUROC & Area Under the Receiver Operating Characteristic Curve \\

AUSGC & Area Under Synthetic Generalization Curve \\

AutoML & Automated Machine Learning \\

BLEU & Bilingual Evaluation Understudy \\

CIDEr & Consensus-based Image Description Evaluation \\

cMMD & conditional Maximum Mean Discrepancy \\

CMS & Centers for Medicare \& Medicaid Services \\

DL & Deep Learning \\

DTW & Dynamic Time Warping \\

DWPro & Dimension-wise Probability \\

DWPre & Dimension-wise Prediction \\

ECG & Electrocardiogram \\

ED & Euclidean Distance \\

EEG & Electroencephalogram \\

EHR & Electronic Health Records \\

GAN &  Generative Adversarial Networks \\ 

HMM & Hidden Markov Model \\

HR & Heart Rate \\

ICU & Intensive Care Unit \\

IS & Inception Score \\

JS-divergence & Jensen–Shannon distance \\

JS-divergence & Jensen–Shannon Divergence\\

KL-divergence & Kullback–Leibler Divergence \\

K-S Test & Kolmogorov–Smirnov Test \\ 

LLM & Large Language Model \\

LPL & Longitudinal Imputation Perplexity \\

LSTM & Long Short-Term Memory \\

ML & Machine Learning \\

MAP & Mean Arterial Pressure \\

MIA & Membership Inference Attack \\

MMD & Maximum Mean Discrepancy \\

MPL & Cross-Modality Imputation Perplexity \\

MSE & Mean Squared Error \\

MVDTW & Multivariate Dynamic Time Warping \\

NLL-Test & Log-Likelihood on the Test Set \\

NODE & Neural Ordinary Differential Equation \\

NSR & Normal Sinus Rhythm \\

PCA & Principal Component Analysis \\

PPG & Photoplethysmogram \\

ROUGE & Recall-Oriented Understudy for Gisting Evaluation \\

SDoH & Social Determinants of Health \\

SHR & Synthetic Health Record \\

t-SNE & t-Distributed Stochastic Neighbor Embedding \\

VAE & Variational Auto-Encoders \\

TRTS & Train on Real, Test on Synthetic \\

TSRTR & Train on Synthetic and Real, Test on Real \\

TSTR & Train on Synthetic, Test on Real \\

\bottomrule

\end{xltabular}

\section{Acknowledgments}
This project received no specific funding. We are thankful to the anonymous reviewers whose valuable insights and constructive feedback have significantly enhanced the quality of this paper. 

\section{Author Contributions}
The authors confirm their contribution to the paper as follows: conceptualization: A.Gh.; database searching: M.L., and M.A.; data curation and analysis: M.L., M.A., and A.Gh.; data interpretation: A.Gh., and M.L.; funding acquisition: not applicable; methodology: M.L., and A.Gh.; project administration: A.Gh.; resources: M.L., and A.Gh.; software: not applicable; supervision: A.Gh.; validation: M.L., A.Gh., and F.P.; visualization: M.L.; writing – original draft: M.L., and A.Gh.; and writing – review \& editing: A.Gh., and M.L. All authors reviewed the results and approved the final version of the manuscript.

%\bibliographystyle{acm}
%\bibliography{references.bib}

\end{document}